\pdfoutput=1
\documentclass[11pt]{article}

\usepackage[]{EMNLP2022}

\usepackage{times}
\usepackage{url}

\usepackage{latexsym}

\usepackage{microtype}

\usepackage{times}
\usepackage{latexsym}
\usepackage{lingmacros}
\usepackage{amsmath}
\usepackage{graphicx}
\usepackage{array, tabularx, caption, boldline}
\usepackage{cellspace}
\usepackage{url}
\usepackage{makecell}
\usepackage{multirow}
\usepackage{xcolor}

\usepackage{subcaption}
\usepackage{booktabs}
\usepackage{comment}
\usepackage{arydshln}
\usepackage{amssymb}

\usepackage{pifont}% http://ctan.org/pkg/pifont
\newcommand{\cmark}{\ding{51}}%
\newcommand{\xmark}{\ding{55}}%

\usepackage{times}
\usepackage{latexsym}

\usepackage[T1]{fontenc}
\usepackage[utf8]{inputenc}

\usepackage{microtype}

\usepackage{inconsolata}

\usepackage{gb4e}
\noautomath
\usepackage{xspace}

\newcommand{\mention}[1]{\textbf{#1}}
\newcommand{\entity}[1]{\textsc{#1}}
\newcommand{\relation}[1]{\texttt{#1}}
\newcommand{\triple}[3]{(\entity{#1}, \relation{#2}, \entity{#3})}
\newcommand{\exref}[1]{(\ref{#1})\xspace}
\newcommand{\ours}{XBE\xspace}
\newcommand{\secref}[1]{(\S\ref{#1})}
\newcommand{\mask}{[M]}

\usepackage{lipsum}

\newcommand\blfootnote[1]{%
  \begingroup
  \renewcommand\thefootnote{}\footnote{#1}%
  \addtocounter{footnote}{-1}%
  \endgroup
}
\usepackage[export]{adjustbox}

\title{Cross-stitching Text and Knowledge Graph Encoders for \\ Distantly Supervised Relation Extraction}

\author{Qin Dai$^\ast$$^1$, Benjamin Heinzerling$^\ast$$^{1,2}$, Kentaro Inui$^{1,2}$ \\ $^1$Tohoku University \qquad $^2$RIKEN AIP \\ \tt qin.dai.b8@tohoku.ac.jp, benjamin.heinzerling@riken.jp \\ \tt kentaro.inui@tohoku.ac.jp}

\begin{document}
\maketitle
\begin{abstract}
Bi-encoder architectures for distantly-supervised relation extraction are designed to make use of the complementary information found in text and knowledge graphs (KG).
However, current architectures suffer from two drawbacks.
They either do not allow any sharing between the text encoder and the KG encoder at all, or, in case of models with KG-to-text attention, only share information in one direction.
Here, we introduce cross-stitch bi-encoders, which allow full interaction between the text encoder and the KG encoder via a cross-stitch mechanism.
The cross-stitch mechanism allows sharing and updating representations between the two encoders at any layer, with the amount of sharing being dynamically controlled via cross-attention-based gates.
Experimental results on two relation extraction benchmarks from two different domains show that enabling full interaction between the two encoders yields strong improvements. \vspace{0.5ex}\\
\includegraphics[width=1em]{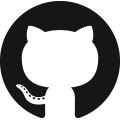} \url{https://github.com/cl-tohoku/xbe}
\blfootnote{\hspace{-1.7em}$\ast$ Equal contribution}

\end{abstract}

\section{Introduction}
\label{sec:intro} 
Identifying semantic relations between textual mentions of entities is a key task for information extraction systems.
For example, consider the sentence:
\begin{exe}
    \ex\label{ex:positive-sentence}\mention{Aspirin} is widely used for short-term treatment of \mention{pain}, fever or colds.
\end{exe}
Assuming an inventory of relations such as \relation{may\_treat} or \relation{founded\_by}, a relation extraction (RE) system should recognize the predicate in \exref{ex:positive-sentence} as an instance of a \relation{may\_treat} relation and extract a knowledge graph (KG) triple like \triple{Aspirin}{may\_treat}{pain}.
RE systems are commonly trained on data obtained via Distant Supervision \citep[DS,][]{mintz2009distant}: Given a KG triple, i.e., a pair of entities and a relation, one assumes that all sentences mentioning both entities express the relation and collects all such sentences as positive examples.
DS allows collecting large amounts of training data, but its assumption is often violated:
\begin{exe}
    \ex\label{ex:false-positive-sentence1}The tumor was remarkably large in size, and \mention{pain} unrelieved by \mention{aspirin}. % TODO: completely unrelated example, different disease and drug or different biomedical example
    \ex\label{ex:false-positive-sentence2}\mention{Elon Musk} fired some \mention{SpaceX} employees who were talking smack about ...  
\end{exe}
Sentence \exref{ex:false-positive-sentence1} is a false positive example of a \relation{may\_treat} relation since it describes a failed treatment.
Sentence \exref{ex:false-positive-sentence2} is, strictly speaking, a false positive of a \relation{founded\_by} relation since this sentence is not about founding companies, but can also be seen as indirect evidence, since founders are often in a position that allows them to fire employees.
We refer to false positive and indirectly relevant examples like \exref{ex:false-positive-sentence1} and \exref{ex:false-positive-sentence2} as \emph{noisy} sentences.

\begin{figure*}[!htb]
    \hspace{-0.8em}
    \begin{tabular}[t]{c|c}
        \begin{tabular}{c}
            \begin{subfigure}[t]{0.43\textwidth}
                \includegraphics[width=\linewidth]{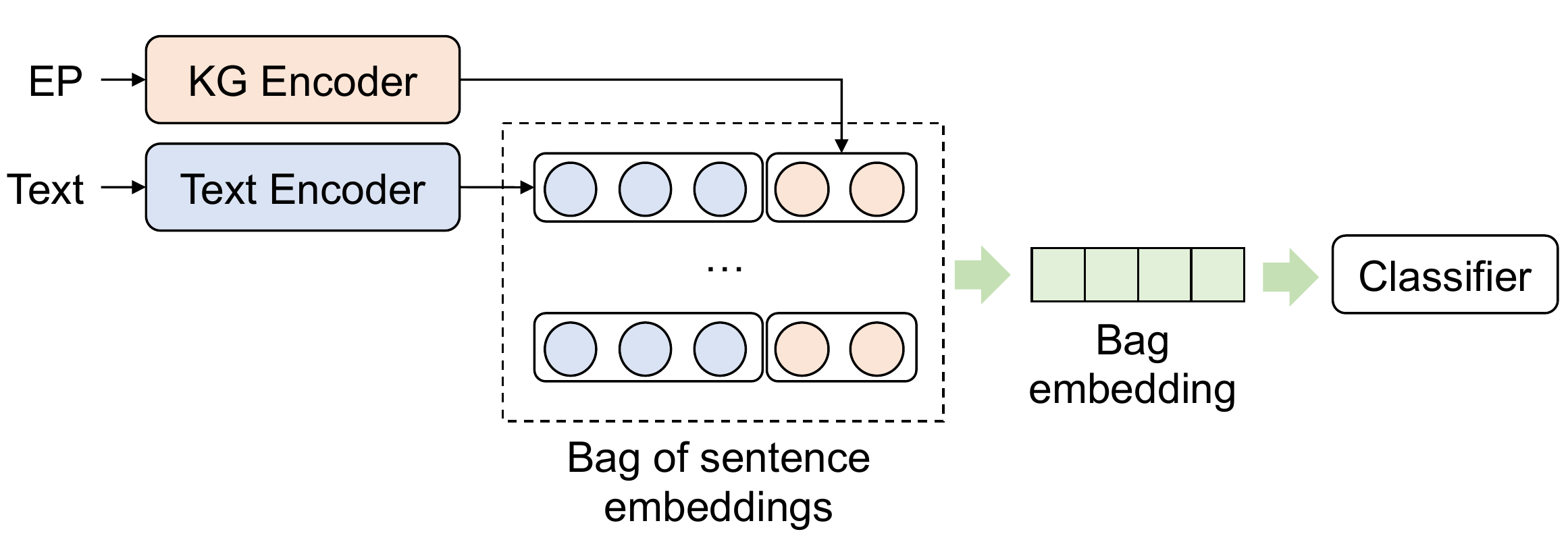}
                \caption{Concatenation of encoder representations}
                \label{fig:kgcat}
                \vspace{3ex}
            \end{subfigure}\\
            \begin{subfigure}[t]{0.43\textwidth}
                \includegraphics[width=\linewidth]{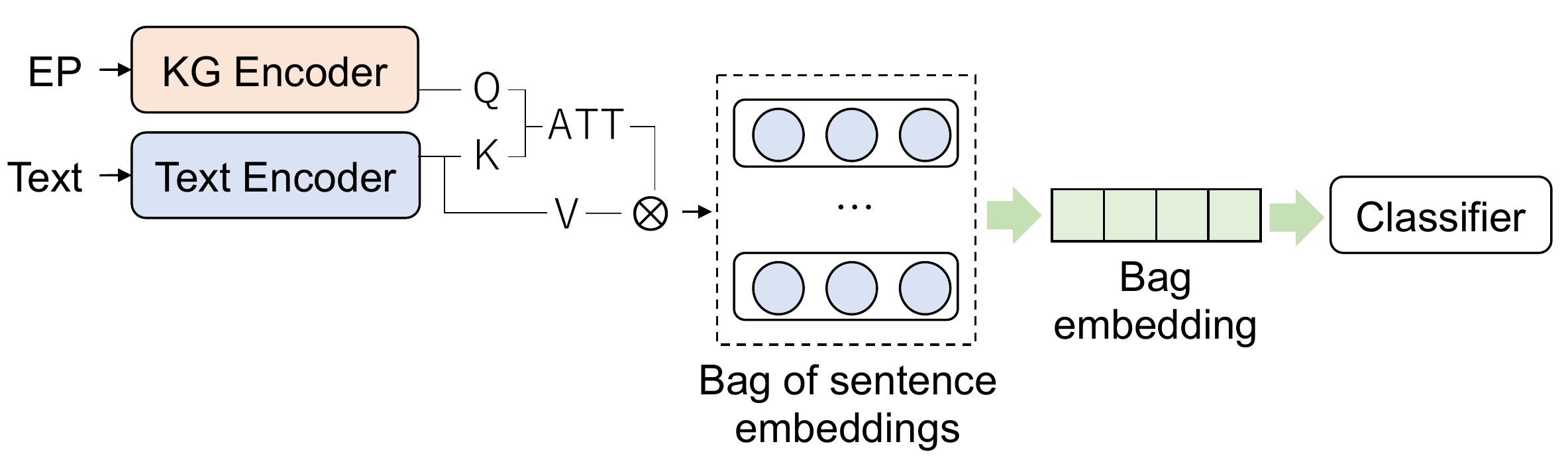}
                \caption{Uni-directional attention from KG-encoder to text-encoder}
                \label{fig:kgatt}
            \end{subfigure}
        \end{tabular}
        \hspace{-0.3em}
        & 
        \hspace{-0.3em}
        \begin{subfigure}{0.53\textwidth}
            \includegraphics[width=\linewidth]{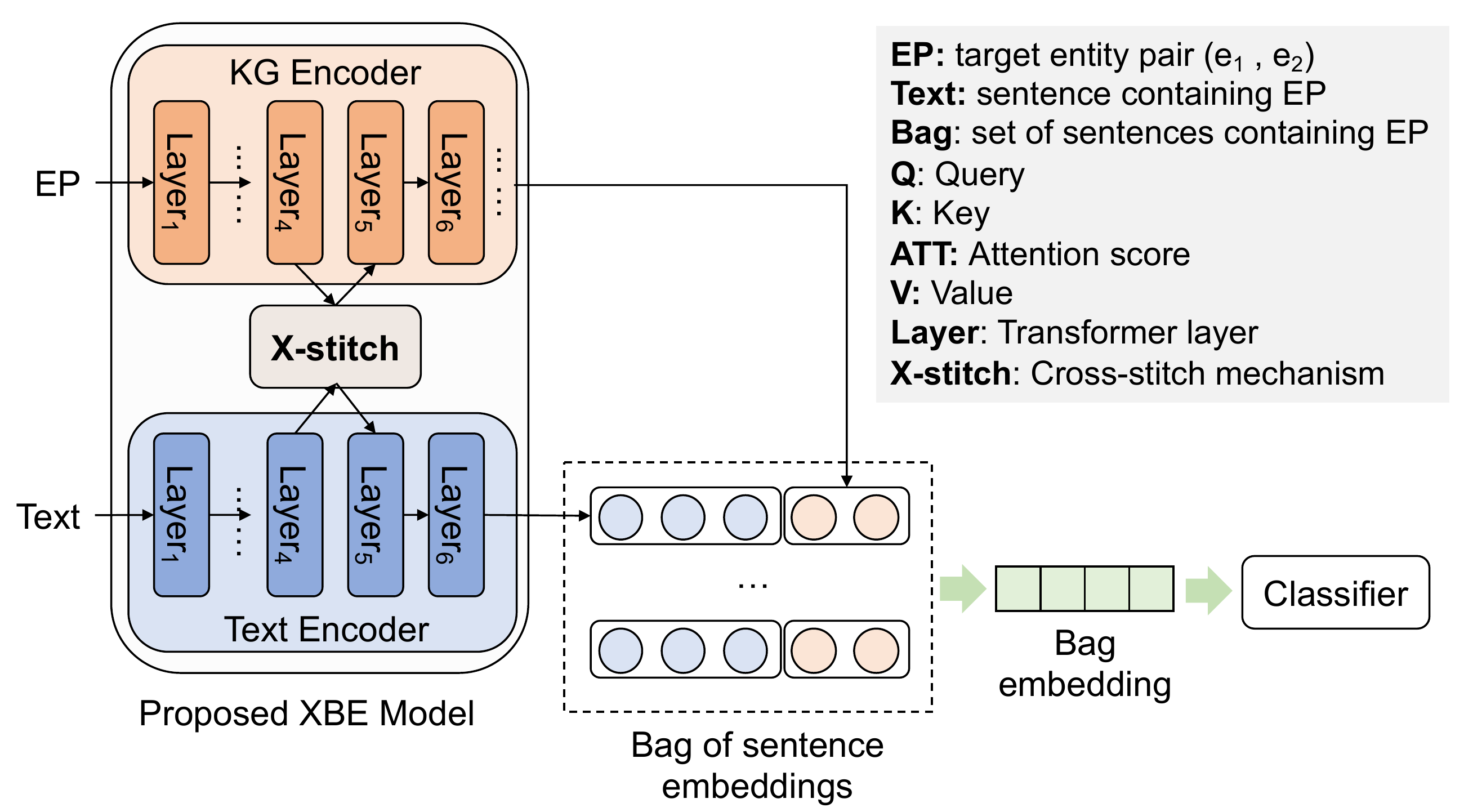}
            \caption{Cross-stitch bi-encoder}
            \label{fig:introxbe}
        \end{subfigure}
    \end{tabular}
  \caption{
  Illustration of existing and proposed bi-encoder architectures for distantly-supervised relation extraction. Simple concatenation of representations (a) does not allow information sharing between text and KG encoders, while KG-to-text attention (b) only allows sharing in one direction. In contrast, our model (c) allows bi-directional information sharing between encoders during the encoding process.}
\label{fig:kgcom}
\end{figure*}

A common approach for dealing with noisy sentences is to use the KG as a complementary source of information.
Models taking this approach are typically implemented as bi-encoders, with one encoder for textual input and one encoder for KG input.
They are trained to rely more on the text encoder when given informative sentences and more on the KG encoder when faced with noisy ones ~\cite{weston2013connecting,han2018neural,zhang2019long,hu2019improving,dai2019incorporating,dai2021two,hu2021knowledge}.
However, current bi-encoder models suffer from drawbacks.
Bi-encoders that encode text and KG separately and then concatenate each encoder's output, as illustrated in Figure~\ref{fig:kgcat} and proposed by \citet{hu2021knowledge}, i.a., cannot share information between the text encoder and the KG encoder during encoding.
In contrast, Bi-encoders whose text encoder can attend to the KG encoder's hidden states, as illustrated in Figure~\ref{fig:kgatt} and proposed by \citet{han2018neural,hu2019improving,zhang2019long}, i.a., do allow information to flow from the KG encoder to the text encoder, but not in the opposite direction.

Here, we propose a cross-stitch bi-encoder (\ours, Figure~\ref{fig:introxbe}) that addresses both of these drawbacks by enabling information sharing between the text encoder and KG encoder at arbitrary layers in both directions.
Intuitively, such a ``full interaction'' between the two encoders is desirable because it is not \textit{a priori} clear at which point in the encoding process an encoder's representation is best-suited for sharing with the other encoder.
Concretely, we equip a bi-encoder with a cross-stitch component \cite{misra2016cross} to enable bi-directional information sharing and employ a gating mechanism based on cross-attention \cite{bahdanau2014neural,vaswani2017attention} to dynamically control the amount of information shared between the text encoder and KG encoder. 
As we will show, allowing bi-directional information sharing during the encoding process, i.e., at intermediate layers, yields considerable performance improvements. 

In summary, our contributions are:%
\begin{itemize}
    \item A bi-encoder architecture that enables full interaction between its encoders: both encoders can share and update information at any layer \secref{sec:model};
    \item An implementation of the proposed architecture for distantly-supervised relation extraction \secref{sec:model-re};
    \item Improvement of performance on two relation extraction benchmarks covering two different domains and achievement of state of the art results on a widely used dataset.\secref{sec:results};
    \item Ablations showing the importance of the components of the proposed architecture \secref{sec:abstudy}.
\end{itemize}

\section{Terminology and Notation}
\label{sec:prelimi}
Throughout this work we use terminology and notation as follows.
We assume access to a domain-specific knowledge graph (KG) which contains fact triples $\mathcal{O}=\{(e_1, r, e_2), ...\}$ consisting of entities $e_1, e_2 \in \mathcal{E}$ and a relation $r \in \mathcal{R}$ that holds between them.
The set of entities $\mathcal{E}$ and the inventory of relations $\mathcal{R}$ are closed and finite.

Given a corpus of entity-linked sentences and KG triples $(e_1^k, r^k, e_2^k)$, distant supervision (DS) yields a bag of sentences $B^k=\{s_1^k,...,s_n^k\}$ where each sentence $s_i^k$ mentions both entities in the pair $(e_1^k, e_2^k)$.
Given the entity pair $(e_1^k, e_2^k)$ and the sentence bag $B^k$, a DS-RE model is trained to predict the KG relation $r^k$.

\section{Cross-stitch Bi-Encoder (XBE)}
\label{sec:model}
The cross-stitch bi-encoder model is designed to enable bidirectional information sharing among its two encoders.
As illustrated in Figure~\ref{fig:introxbe}, it consists of a text encoder, a KG encoder, and a cross-stitch component controlled by cross-attention.
The following subsections describe these components.

\subsection{Bi-Encoder}
To obtain representations of inputs belonging to the two different modalities in DS-RE, we employ a bi-encoder architecture consisting of one encoder for textual inputs and one encoder for KG triples.
While the cross-stitch component is agnostic to the type of encoder, we use pre-trained Transformer models \cite{vaswani2017attention} for both text and KG.

The \textbf{Text Encoder} takes a sentence $s_i^k$ containing a sequence of $N$ tokens $(tok_1,...,tok_N)$ as input and produces $L_T$ layers of $d_T$-dimensional contextualized representations $S_i \in \mathbb{R}^{N \times d_T}$, $1 \leqslant i \leqslant L_T$.
We construct a fixed-length representation of the sentence $s_i^k$ mentioning the entity pair $(e_1^k, e_2^k)$ by concatenating
the embeddings of the head and tail entities $h_{e1}$ and $h_{e2}$ obtained from the last layer $S_{L_T}$ via the method described in \citet{peng2020learning}, as well as the mean- and max-pooled token representations $h_{mean}$ and $h_{max}$ obtained from pooling over the last encoder layer $S_{L_T}$.
That is, the final representation of the input sentence $s_i^k$ is $\mathbf{s}_i^k = [h_{e1};h_{e2};h_{mean};h_{max}]$, where $;$ denotes vector concatenation.

The \textbf{KG Encoder} takes a KG triple $(e_1, r, e_2)$ as input and generates $T_K$~layers of $d_K$-dimensional contextualized representations 
$T_i \in \mathbb{R}^{3 \times d_K}$, $1 \leqslant i \leqslant L_K$. Then $x_{e_1} \in \mathbb{R}^{d_K}$, $x_r \in \mathbb{R}^{d_K}$ and $x_{e_2} \in \mathbb{R}^{d_K}$ from the last layer $T_{L_K}$ are used as the embeddings of the head entity $e_1$, relation $r$ and tail entity $e_2$ respectively.
The KG encoder's vocabulary $\mathcal{V}$ is formed by the union of all entities $\mathcal{E}$ and relations $\mathcal{R}$, as well as a mask token \text{\mask}, i.e., $\mathcal{V} = \mathcal{R} \cup \mathcal{E} \cup \{\text{\mask}\}$.
For simplicity we assume that the text and KG encoder representations have the same dimensionality $d$, that is, we set $d = d_K = d_L$, although this is not required by the model architecture.

\subsection{Cross-stitch (X-stitch)}
\label{sec:crst}
\begin{figure}[t]
\centering
\includegraphics[width=\columnwidth]{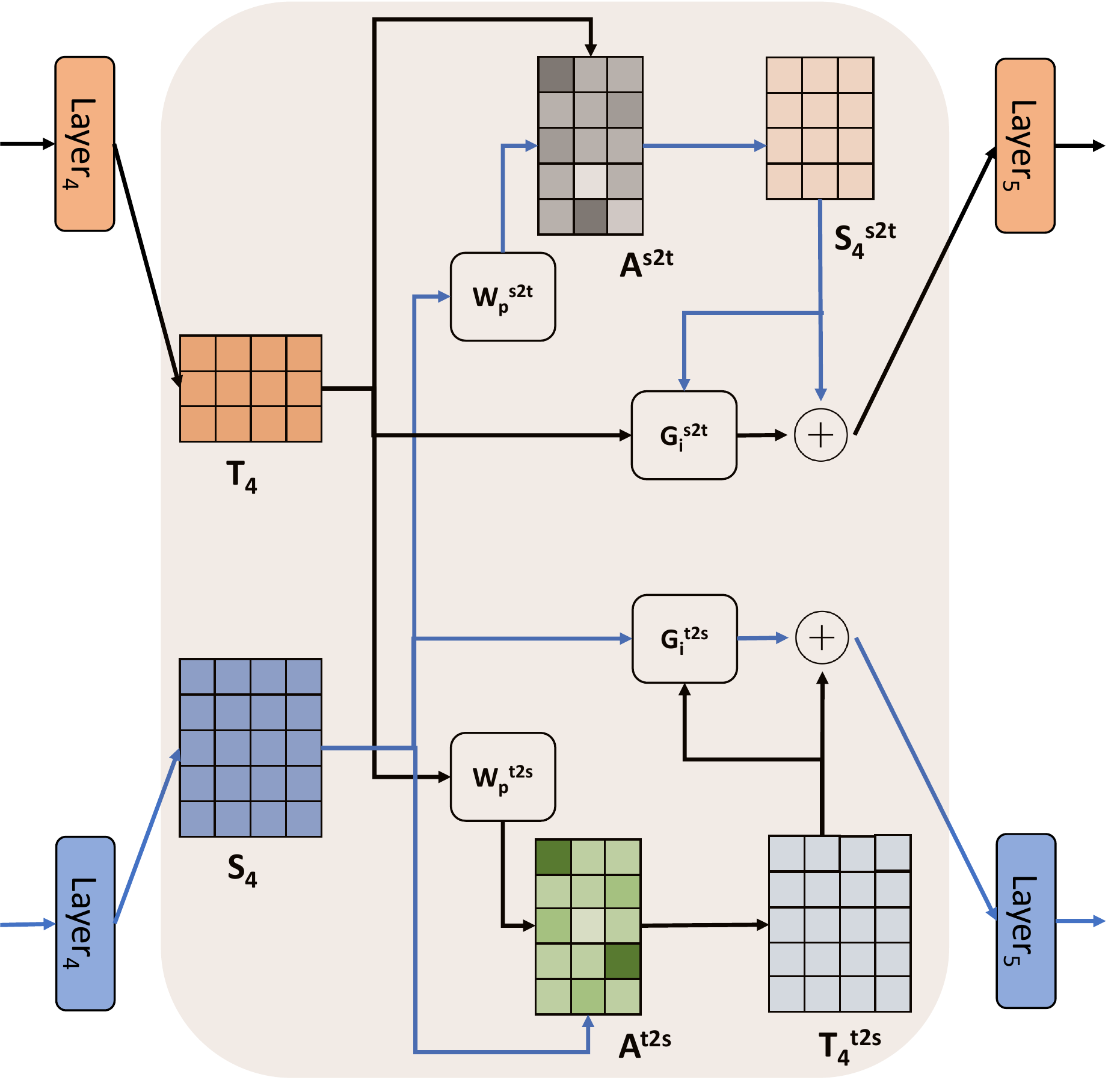}
\caption{Illustration of the cross-stitch mechanism in combination with cross-attention. See~\S\ref{sec:crst} for notation.}
\label{fig:crst}
\end{figure}
To enable bi-directional information sharing between the two encoders, we employ a cross-stitch\footnote{%
For brevity, we use \emph{X-stitch} in tables and figures.} mechanism based on \citet{misra2016cross}.
The mechanism operates by mixing and updating intermediate representations of the bi-encoder.
We dynamically control the amount of mixing via gates based on cross-attention (Figure~\ref{fig:crst}).
More formally, our cross-stitch variant operates as follows.
Given a sentence $s=(tok_1,...,tok_N)$ and corresponding KG triple $t=(e_1, r, e_2)$, the text encoder generates sentence representations $S_i \in \mathbb{R}^{N \times d}$and the KG encoder triple representations $T_i \in \mathbb{R}^{3 \times d}$.
We then compute cross-attentions $A$ in two directions, triple-to-sentence ($t2s$) and sentence-to-triple ($s2t$), via Equations~\ref{eq:t2s} and \ref{eq:s2t},
\begin{align}\label{eq:t2s}
A^{t2s} = \textrm{softmax}_{\textrm{column}}((W_p^{t2s} \cdot T_i) \cdot S_i)
\end{align}
\begin{align}\label{eq:s2t}
A^{s2t} = \textrm{softmax}_{\textrm{row}}(S_i \cdot (W_p^{s2t} \cdot T_i)^T)
\end{align}
where, $W_p^{s2t} \in \mathbb{R}^{d \times d}$ and $W_p^{t2s} \in \mathbb{R}^{d \times d}$ denote trainable linear transformations.
The triple-to-sentence attention $A^{t2s}$ represents the weight of the embedding of each token in triple $t$ that will be used to update the sentence representation $S_i$:
\begin{align}\label{eq:compt2s}
T_i^{t2s} = W_{g2}^{t2s} \cdot ReLU(W_{g1}^{t2s} \cdot (A^{t2s} \cdot T_i^T))
\end{align}
where $W_{g1}^{t2s} \in \mathbb{R}^{d' \times d}$ and $W_{g2}^{t2s} \in \mathbb{R}^{d \times d'}$ are trainable parameters.
Next, a gating mechanism determines the degree to which the original textual representation $S_i$ will contribute to the new hidden state of the text encoder:
\begin{align}\label{eq:gates}
\mathbf{G}_i^{t2s} = \sigma (T_i^{t2s})
\end{align}
where, $\sigma$ denotes the logistic sigmoid function.
We then update the hidden state of the text encoder at layer $i$ by interpolating its original hidden state $S_{i}$ with the triple representation $T_i^{t2s}$:

\begin{align}\label{eq:news}
S_i' = \mathbf{G}_i^{t2s} \cdot S_i + \lambda_t \cdot T_i^{t2s} 
\end{align}
Information sharing in the sentence-to-triple direction is performed analogously:
\begin{align}\label{eq:comps2t}
S_i^{s2t} = W_{g2}^{s2t} \cdot ReLU(W_{g1}^{s2t} \cdot ((A^{s2t})^T \cdot S_i))
\end{align}
\begin{align}\label{eq:gatet}
\mathbf{G}_i^{s2t} = \sigma (S_i^{s2t})
\end{align}

\begin{align}
T_i' = \mathbf{G}_i^{s2t} \cdot T_i  + \lambda_s \cdot S_i^{s2t}
\end{align}
\label{eq:newt}
where $\lambda_t$ and $\lambda_s$ are weight hyperparameters. Having devised a general architecture for text-KG bi-encoders, we now turn to implementing this architecture for distantly supervised relation extraction.

\section{XBE for Relation Extraction}
\label{sec:model-re}

In distantly supervised relation extraction, the automatically collected data consists of a set of sentence bags $\{B^1, ..., B^n\}$ and set of corresponding KG triples  $\{(e_1^1,r^1,e_2^1), ..., (e_1^n,r^n,e_2^n)\}$.
To create training instances, we mask the relation in the KG triples $\{(e_1^1,\text{\mask},e_2^1), ..., (e_1^n,\text{\mask},e_2^n)\}$ and provide these masked triples as input to the KG encoder, while the text encoder receives one sentence from the corresponding sentence bag.
If the sentence bag contains $k$ sentences, we pair each sentence with the same KG triple and run the bi-encoder for each pairing, i.e., $k$ times, to obtain a sentence bag representation.
During training, the loss of the model is calculated via  Equations~\ref{eq:allloss}, \ref{eq:reloss} and \ref{eq:lploss},
\begin{align}\label{eq:allloss}
L = L_{RE} + w \cdot L_{KG}
\end{align}
\begin{align}\label{eq:reloss}
L_{RE}=-\sum^n_{k=1}\sum^{|B^k|}_{i=1}\log P(r^k|[\mathbf{s}_i^k;\mathbf{r}_{ht};x_{e_1^k};x_{e_2^k}])
\end{align}
\begin{align}\label{eq:lploss}
L_{KG}=-\sum^n_{k=1}\log g((e_1^k, \text{\mask}, e_2^k))
\end{align}
where $w \in (0, 1]$ is a weight hyperparameter, $P(x)$ is the predicted probability of the target relation over a set of predefined relations, $\mathbf{r}_{ht}$ is an additional KG feature vector obtained from a pre-trained KG completion model such as TransE~\cite{bordes2013translating}, $L_{KG}$ is the loss of KG relation prediction and $g(x)$ outputs the predicted probability of the masked token over the vocabulary $\mathcal{V}$ based on the embedding $x_{\text{\mask}}$ from the KG encoder.

During inference, we follow \citet{hu2021knowledge} and use the mean of sentence embeddings as the bag embedding:
\begin{align}\label{eq:infer}
P(r^k|B^k)=(\sum^{|B^k|}_{i=1} P(r^k|[\mathbf{s}_i^k;\mathbf{r}_{ht};x_{e_{1}^k};x_{e_{2}^k}]))/|B^k|
\end{align}

As our bi-encoder consists of two transformer-based encoders, we make use of pre-training for each modality.
For the text encoder, we employ an off-the-shelf model, as detailed in the next section.
The KG encoder is pre-trained on a set of KG triples via a relation prediction task.
Specifically, given a relation masked triple $(e_1, \text{\mask}, e_2)$, the KG encoder is asked to predict the masked symbolic token and pre-trained via the loss given by Equation~\ref{eq:lploss}. 
%
%\begin{align}\label{eq:kgloss}
%L_{KG}=-\sum^M_{i=1}\log g((e_1^i, \text{\mask}, e_2^i))
%\end{align}
%
%where $M$ is the total number of masked KG triples and $g(x)$ outputs the predicted probability of the masked token over the vocabulary $\mathcal{V}$ based on the embedding $x_{\text{\mask}}$ from the KG encoder.

\section{Experiments}

\subsection{Data}
\label{sec:data}
We evaluate our model on the biomedical dataset introduced by \citet{dai2021two} (hereafter: Medline21) and the NYT10 dataset~\cite{riedel2010modeling}
Statistics for both datasets are summarized in Table~\ref{tab:kb_sta}.

\textbf{Medline21}. This dataset was created by aligning the biomedical knowledge graph UMLS\footnote{\url{https://www.nlm.nih.gov/research/umls/}} with the Medline corpus,  a collection of biomedical abstracts.
Both resources are published by the U.S. National Library of Medicine\footnote{\url{https://www.nlm.nih.gov/}}.
A state-of-the-art UMLS Named Entity Recognizer, ScispaCy~\cite{neumann-etal-2019-scispacy}, is applied to identify UMLS entity mentions in the Medline corpus.
The sentences until the year 2008 are used for training and the ones from the year 2009 $\sim$ 2018 are used for testing.
Following \cite{han2018neural}, \newcite{dai2021two} also provided a subset of UMLS in the dataset, which consists of $582,686$ KG triples. We use the set of triples to train the KG encoder.

\textbf{NYT10}. This dataset was created by aligning Freebase relational facts with the New York Times Corpus. Sentences from the year 2005 $\sim$ 2006 are used for training and the sentences from 2007 are used for testing. The NYT10 dataset has been used widely for relation extraction \cite{lin2016neural,ji2017distant,du2018multi,jat2018improving,han2018neural,han2018hierarchical,vashishth2018reside,ye2019distant,hu2019improving,alt2019fine,sun2019leveraging,li2020self,hu2021knowledge,dai2021two}. In order to leverage a KG for DS-RE on NYT10, \citet{han2018neural} extended the dataset with FB60K, which is a KG containing $335,350$ triples. Following \cite{hu2019improving,han2018neural,hu2021knowledge}, we use FB60K to train the KG encoder for DS-RE.
\begin{table}[t]
\centering
\scalebox{0.8}{
\begin{tabular}{c|c|c|c|c}
\hline
&\textbf{\#R} & \textbf{\#EP}& \textbf{\#Related EP} & \textbf{\#Sentence} \\
\hline
\hline
Medline21 & 40 & \makecell{100,549 /\\ 21,081} & \makecell{10,936 /\\ 1,804} & \makecell{165,692 /\\ 28,912} \\ \hline
NYT10 & 53 & \makecell{281,270 /\\ 96,678} & \makecell{18,252 /\\ 1,950} & \makecell{522,611 /\\  172,448} \\ \hline
\end{tabular}
}
\caption{Statistics of datasets in this work, where \textbf{R} and \textbf{EP} stand for the target Relation and Entity Pair, $\#_1$/$\#_2$ represent the number of training and testing data respectively.}
\label{tab:kb_sta}
\end{table}

\subsection{Settings}
\label{sec:settings}
Following the conventional settings of DS-RE \cite[see, e.g.,][] {lin2016neural}, we conduct a held-out evaluation, in which  models are evaluated by comparing the fact triple identified from a bag of sentences $S_r$ with the corresponding KG triple. Further following evaluation methods of previous work, we draw Precision-Recall curves and report the Area Under Curve (AUC), as well as Precision@N (P@N) scores, which give the percentage of correct triples among the top N ranked predictions.
In addition, as done by \citet{hu2021knowledge}, the text encoder (\S\ref{sec:model}) for experiments on NYT10 is initialized with the pre-trained weights from the \texttt{bert-base-uncased} variant of BERT \cite{devlin2018bert}.
The text encoder for Medline21 is initialized with BioBERT~\cite{lee2020biobert} and the KG encoder (\S\ref{sec:model}) is pre-trained using each dataset's corresponding KG, as mentioned above.

%The parameter settings of our experiments are detailed in the supplementary material.

\begin{figure*}[t]
\centering
\begin{minipage}{.5\textwidth}
  \centering
  \includegraphics[width=\columnwidth]{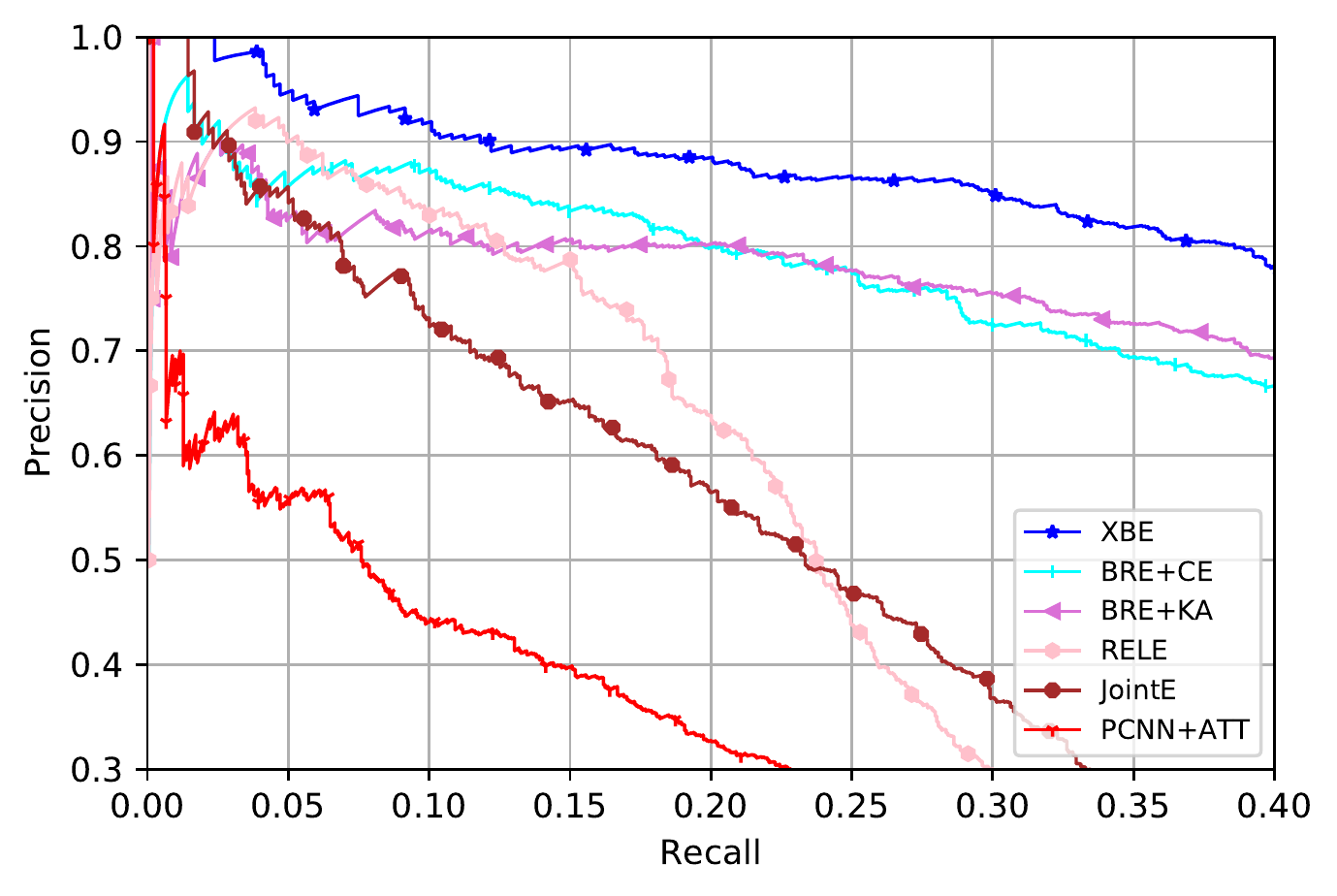}
  \captionof{figure}{PR curves on Medline21.}
  \label{fig:prc_bio}
\end{minipage}%
\begin{minipage}{.5\textwidth}
  \centering
  \includegraphics[width=\columnwidth]{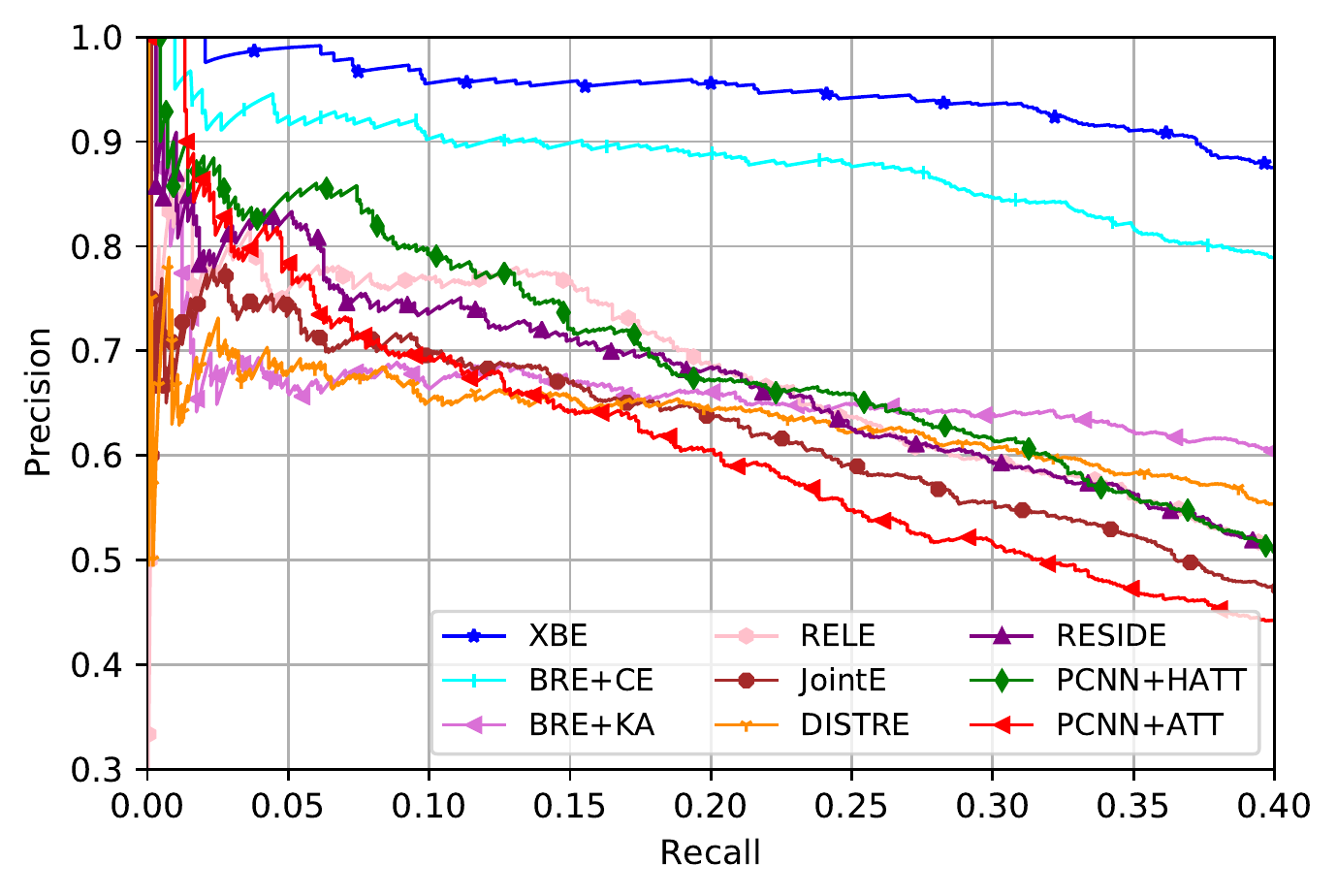}
  \captionof{figure}{PR curves on NYT10.}
  \label{fig:prc_nyt}
\end{minipage}
\end{figure*}

\begin{table*}[t]
\centering
\scalebox{0.78}{
\begin{tabular}{lccccccccccccccc}
\hline
&\multicolumn{5}{c}{\textbf{Medline21}} &\multicolumn{7}{c}{\textbf{NYT10}}\\\cmidrule(lr){2-6}\cmidrule(lr){7-13}
\textbf{Model} & AUC & P@0.3k & P@0.5k & P@1k & P@2k & AUC & P@0.1k & P@0.2k & P@0.3k & P@0.5k & P@1k & P@2k\\\cmidrule(lr){1-6}\cmidrule(lr){7-13}\cmidrule(lr){1-6} \cmidrule(lr){7-13}
PCNN+ATT & 17.8$^\ast$ & 48.3$^\ast$ & 43.2$^\ast$ & 34.3$^\ast$ & 25.2$^\ast$ & 34.1$^\dag$ & 73.0$^\dag$ & 68.0$^\dag$ & 67.0$^\dag$ & 63.6$^\dag$ & 53.3$^\dag$ & 40.0$^\dag$ \\\cmidrule(lr){1-6}\cmidrule(lr){7-13}
PCNN+HATT & - & - & - & - & - & 42.0$^\ddag$ & 81.0$^\ddag$ & 79.5$^\ddag$ & 75.7$^\ddag$ & 68.0$^\ddag$ & 58.6$^\ddag$ & 42.1$^\ddag$ \\\cmidrule(lr){1-6}\cmidrule(lr){7-13}
RESIDE & - & - & - & - & - & 41.5$^\dag$ & 81.8$^\dag$ & 75.4$^\dag$ & 74.3$^\dag$ & 69.7$^\dag$ & 59.3$^\dag$ & 45.0$^\dag$ \\\cmidrule(lr){1-6}\cmidrule(lr){7-13}
DISTRE & - & - & - & - & - & 42.2$^\dag$ & 68.0$^\dag$ & 67.0$^\dag$ & 65.3$^\dag$ & 65.0$^\dag$ & 60.2$^\dag$ & 47.9$^\dag$ \\\hline\hline
JointE & 26.3$^\star$ & 70.0$^\star$ & 61.4$^\star$ & 46.4$^\star$ & 30.0$^\star$ & 38.5$^\star$ & 74.0$^\star$ & 71.5$^\star$ & 69.0$^\star$ & 65.4$^\star$ & 55.9$^\star$ & 43.6$^\star$\\\cmidrule(lr){1-6}\cmidrule(lr){7-13}
RELE & 25.6$^\star$ & 78.7$^\star$ & 66.8$^\star$ & 44.7$^\star$ & 27.5$^\star$ & 40.5$^\star$ & 79.0$^\star$ & 77.0$^\star$ & 77.0$^\star$ & 71.2$^\star$ & 59.3$^\star$ & 44.7$^\star$ \\\cmidrule(lr){1-6}\cmidrule(lr){7-13}
BRE+KA & 50.3$^\star$ & 79.7$^\star$ & 79.2$^\star$ & 70.3$^\star$ & 51.2$^\star$ & 48.8$^\star$ & 68.0$^\star$ & 68.0$^\star$ & 67.0$^\star$ & 66.0$^\star$ & 63.7$^\star$ & 52.4$^\star$ \\\cmidrule(lr){1-6}\cmidrule(lr){7-13}
BRE+CE & 55.3$^\star$ & 84.0$^\star$ & 79.4$^\star$ & 67.7$^\star$ & 53.8$^\star$ & 63.2$^\ddag$ & 92.0$^\ddag$ & 92.0$^\ddag$ & 90.0$^\ddag$ & 88.0$^\ddag$ & 78.7$^\ddag$ & 58.7$^\ddag$ \\\cmidrule(lr){1-6}\cmidrule(lr){7-13}
XBE & \textbf{61.9} & \textbf{89.3} & \textbf{86.4} & \textbf{76.1} & \textbf{56.1} & \textbf{70.5} & \textbf{99.0} & \textbf{96.0} & \textbf{95.6} & \textbf{94.4} & \textbf{85.8} & \textbf{63.2} \\\hline
\end{tabular}
}
\caption{P@N and AUC on Medline21 and NYT10 datasets (k=1000), where \dag represents that these results are quoted from \cite{alt2019fine}, \ddag indicates the results using the pre-trained model, $\star$ indicates the results are obtained by re-running corresponding codes and $\ast$ indicates using the OpenNRE~\cite{han-etal-2019-opennre} implementation.}
\label{tab:patn_bio_nyt}
\end{table*}

\subsection{Baseline Models}
\label{sec:baselines}
To demonstrate the effectiveness of the proposed model, we compare to the following baselines.
Baselines were selected because they are the closest models in terms of integrating KG with text for DS-RE and/or because they achieve competitive or state-of-the-art performance on the datasets used in our evaluation. 
\begin{itemize}
    \item \textbf{JointE}~\cite{han2018neural}: A joint model for KG embedding and RE, where the KG embedding is utilized for attention calculation over a sentence bag, as shown in Figure~\ref{fig:kgatt}.
    \item \textbf{RELE}~\cite{hu2019improving}: A multi-layer attention-based model, which makes use of KG embeddings and entity descriptions for DS-RE. 
    \item \textbf{BRE+KA}~\cite{hu2021knowledge}: A version of the JointE model that integrates BERT.
    \item \textbf{BRE+CE}~\cite{hu2021knowledge}: A BERT and KG embedding based model, where BERT output and the KG triple embedding are concatenated as a feature vector for DS-RE, as shown in Figure~\ref{fig:kgcat}.
\end{itemize}
To collect AUC results and draw Precision-Recall curves, we use pre-trained models where possible or carefully run published implementations using suggested hyperparameters\footnote{For hyperparameter selection on Medline21, we use a $30$\% random split of the training set.} from the original papers if no pre-trained model is publicly available. See the supplementary material for training details.

In addition to the models above, we select the following baselines for further comparison. 
\begin{itemize}
    \item \textbf{PCNN+ATT}~\cite{lin2016neural} A CNN-based model with a relation embedding attention mechanism.
    \item \textbf{PCNN+HATT}~\cite{han2018hierarchical} A CNN-based model with a relation hierarchy attention mechanism.
    \item \textbf{RESIDE}~\cite{vashishth2018reside} A Bi-GRU-based model which makes use of relevant side information (e.g., syntactic information), which is encoded via a Graph Convolution Network.
    \item \textbf{DISTRE}~\cite{alt2019fine} A Generative Pre-trained Transformer model with a relation embedding attention mechanism. 
\end{itemize}

\subsection{Results}
\label{sec:results}

The Precision-Recall (PR) curves of each model on Medline21 and NYT10 datasets are shown in Figure~\ref{fig:prc_bio} and Figure~\ref{fig:prc_nyt}, respectively.
We make two main observations: (1) Among the compared models, BRE+KA and BRE+CE, are strong baselines because they significantly outperform other state-of-the-art models especially when the recall is greater than $0.25$, demonstrating the benefit of combining a pre-trained language model (here: BERT) and a KG for DS-RE.
(2) The proposed XBE model outperforms all baselines and achieves the highest precision over the entire recall range on both datasets.
Table~\ref{tab:patn_bio_nyt} further presents more detailed results in terms of AUC and P@N, which shows improved performance of XBE in all testing metrics. In particular, XBE achieves a new state-of-the-art on the commonly used NYT10 dataset.

Since the underlying resources, namely the pre-trained language model and the KG are the same as those used by the best baseline models, we take this strong performance as evidence that the proposed model can make better use of the combination of KG and text.
This in turn, we hypothesize, is due to the fact that our proposed model can realize encoder layer level communication between KG and text representations.
In the next section we conduct an ablation study to verify this hypothesis.

%\textbf{Comparison with Various State-of-the-art Baselines on NYT10}. To demonstrate the effectiveness of our XBE model, we also compare it against the following baselines on NYT10 dataset: Mintz~\cite{mintz2009distant}, MultiR~\cite{hoffmann2011knowledge}, MIMLRE~\cite{surdeanu2012multi}, PCNN~\cite{zeng2015distant}, PCNN+ATT~\cite{lin2016neural}, BGWA~\cite{jat2018improving}, PCNN+HATT~\cite{han2018hierarchical}, RESIDE~\cite{vashishth2018reside}, DISTRE~\cite{alt2019fine}, Sent+KG-path~\cite{dai2019incorporating}, UGDSRE~\cite{dai2021two} and BRE+CE~\cite{hu2021knowledge}.
%The results shown in Figure~\ref{fig:prc_stnyt} and Table~\ref{tab:patn_stnyt} indicate that: (1) the selected model, ``BRE+CE'', is a strong baseline because it significantly outperforms other state-of-the-art models; and (2) XBE can effectively take advantage of the integration of KG and text for DS-RE, because it beats the strong baseline and achieves a new state-of-the-art result on the commonly used DS-RE dataset.

\subsection{Ablation Study}
\label{sec:abstudy}
\begin{table}[t]
\centering
\scalebox{0.9}{
\begin{tabular}{lcccc}
\hline
&\multicolumn{2}{c}{\textbf{Medline21}} &\multicolumn{2}{c}{\textbf{NYT10}}\\\cmidrule(lr){2-3}\cmidrule(lr){4-5}
\textbf{Model} & AUC & P@2k & AUC & P@2k\\\cmidrule(lr){1-3}\cmidrule(lr){4-5}
\ours & \textbf{61.9} & \textbf{56.1} & \textbf{70.5} & \textbf{63.2} \\\cmidrule(lr){1-3}\cmidrule(lr){4-5}
\makecell[r]{- X-stitch} & 58.7 & 53.3 & 68.3 & 61.3 \\\cmidrule(lr){1-3}\cmidrule(lr){4-5}
\makecell[r]{- KG enc.} & 55.7 & 53.8 & 61.5 & 56.9 \\\cmidrule(lr){1-3}\cmidrule(lr){4-5}
\makecell[r]{- text enc.} & 39.8 & 41.1 & 55.9 & 55.1 \\\hline
\end{tabular}
}
\caption{Performance comparison of XBE with different ablated components (non-cumulative) on Medline21 and NYT10 datasets (k=1000).}
\label{tab:ablation}
\end{table}
\begin{table}[t]
\centering
\scalebox{0.9}{
\begin{tabular}{lcccc}
\hline
&\multicolumn{2}{c}{\textbf{Medline21}} &\multicolumn{2}{c}{\textbf{NYT10}}\\\cmidrule(lr){2-3}\cmidrule(lr){4-5}
\textbf{Model} & AUC & P@2k & AUC & P@2k\\\cmidrule(lr){1-3}\cmidrule(lr){4-5}
\ours & \textbf{61.9} & \textbf{56.1} & \textbf{70.5} & \textbf{63.2} \\\cmidrule(lr){1-3}\cmidrule(lr){4-5}
\makecell[r]{- Pre-KG enc.} & 55.8 & 52.8 & 63.7 & 60.3 \\\cmidrule(lr){1-3}\cmidrule(lr){4-5}
\makecell[r]{- Joint-KG enc.} & 49.7 & 49.6 & 63.0 & 59.1 
\\\hline
\end{tabular}
}
\caption{Performance comparison of XBE trained under different conditions (non-cumulative) on Medline21 and NYT10 datasets (k=1000).}
\label{tab:ablationkg}
\end{table}
\begin{figure*}[t]
\centering
\includegraphics[width=14.0cm]{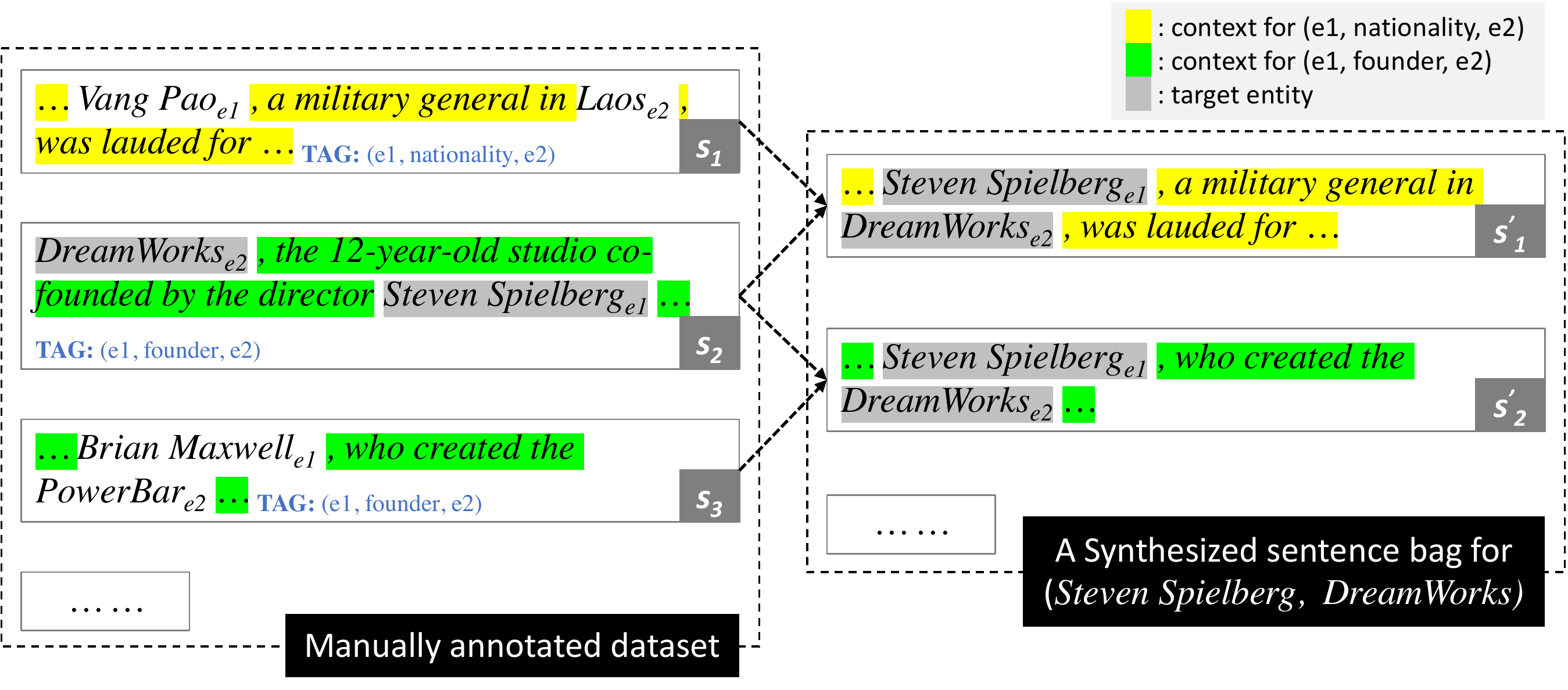}
\caption{Process of creating a synthesized sentence bag, in which a valid sentence (e.g., $s'_2$) is created by the combination of a target entity pair (e.g., (Steven Spielberg, DreamWorks)) and the context representing their relation (e.g., \emph{founder}), while a noisy one (e.g., $s'_1$) is done by the combination of the target entity pair and a random context representing different relation (e.g., \emph{nationality}).}
\label{fig:synsent}
\end{figure*}
\begin{figure}[t]
\centering
\includegraphics[width=\columnwidth]{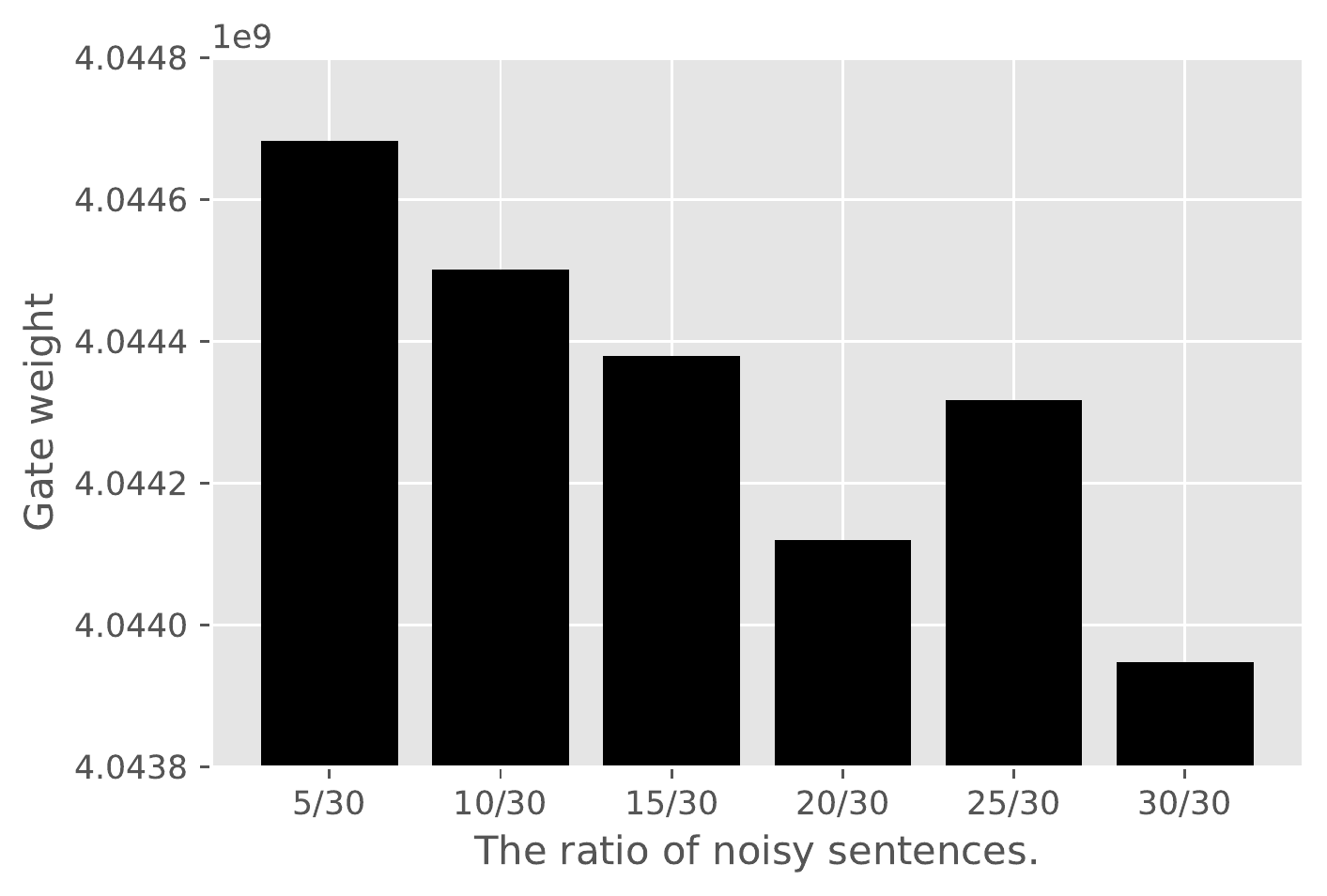}
\caption{The sum of gate weights w.r.t different noise ratio. The x axis denotes the noise ratio of a set of synthesized data, where $\#_1$/$\#_2$ represents the number of noisy and all sentences in a synthesized sentence bag respectively. The y axis denotes the sum of gate weights over an entire set.}
\label{fig:noisegate}
\end{figure}
We first ablate the three main model components in order to assess the contribution to overall performance.
Results are shown in Table~\ref{tab:ablation}, where ``- X-stitch'' is the model without the cross-stitch  mechanism, ``- KG enc.'' denotes removing the KG encoder, and ``- text enc.'' removing the text encoder.
We observe that performance drops for all ablations, indicating that each component is important for the model when performing DS-RE.
While the impact of ablating the text encoder is by far the largest, removing the cross-stitch component or the KG encoder results in performance that is comparable to the performance of the strongest baseline, BRE+CE, on both datasets.
This suggests that these two components, i.e., the KG encoder and the cross-stitch mechanism allowing sharing of information between the text and KG encoder, are what enables our model to improve over BRE+CE.

As described in~\S\ref{sec:model-re}, we pre-train the KG encoder via a relation prediction task before fine-tuning the XBE model end-to-end on a DS-RE dataset.
In order to measure the effect of KG encoder pre-training, we compare with a setup in which the KG encoder is not pre-trained but initialized randomly instead.
In addition, since our proposed XBE model facilitates joint training of the KG encoder and text encoder, we also compare to a setting in which the pre-trained KG encoder is frozen, i.e., not updated during training on the two DS-RE datasets. 
The results of these KG-encoder ablations are shown in Table~\ref{tab:ablationkg}, where ``- Pre-KG enc.'' denotes the random initialization of the KG encoder and ``- Joint-KG enc.'' is the model with a pre-trained, frozen KG encoder.
We observe that performance decreases both without pre-training of the KG encoder and when we freeze the KG encoder while fine-tuning XBE.
That is, performance gains not only stem from employing a pre-trained KG encoder but also from the effective joint training of both the KG encoder and the text encoder.

\begin{table*}[t]
\centering
\scalebox{0.8}{
\begin{tabular}{c|l|l|c|c|c}
\hline
Bag &\textbf{Sentence} & Target Relation &\textbf{XBE}& \makecell{XBE \\- X-stitch} & BRE+CE \\
\hline
\hline
B1 & \makecell[l]{... is released from prison into an unrecognizable 1980s \textbf{Paris}, \\ and ... in which the New Wave aesthetic reaches some \\ sort of terminal point, as \textbf{Jean-Pierre} ...} & \makecell{/people/person/\\place\_lived} & \cmark & \xmark & \xmark\\ \hline
B2 & \makecell[l]{... an online advertising technology company in \textbf{san francisco} \\ called \textbf{zedo}, ...} & \makecell{/business/company/\\place\_founded} & \cmark & \xmark & \xmark\\ \hline\hline
B3 & \makecell[l]{... CagA promoted the underglycosylation of IgA1 ,\\ which at least partly attributed to the downregulation \\ of \textbf{C3812673\#ent} ( C1GALT1 ) and its \textbf{C1332924\#ent}  ...} & \makecell{gene\_product\_ \\ encoded\_by\_gene} & \cmark & \xmark & \xmark\\ \hline
B4 & \makecell[l]{Although existing recommendations for the care of \\ patients with \textbf{C0017205\#ent} in effect , \\ unique characteristics of \textbf{C3272698\#ent} \\ require additional investigation and monitoring .} & \makecell{may\_be\_treated\_by} & \cmark & \xmark & \xmark\\ \hline
\end{tabular}
}
\caption{Qualitative results. Each bag contains one sentence, \cmark (or \xmark) represents the correct (or incorrect) prediction of the target relation.}
\label{tab:casestudy}
\vspace{3ex}
\end{table*}
\begin{figure*}
\centering
\begin{subfigure}{15.0cm}
  \centering
  \includegraphics[width=14.0cm,left]{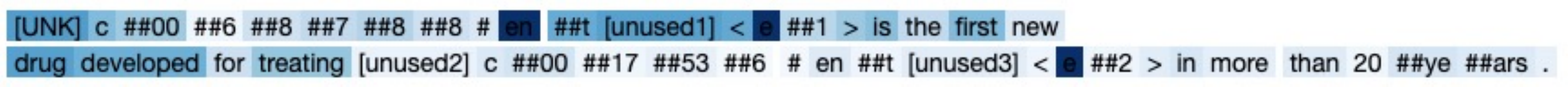}
  \caption{}
  \label{fig:bioatt}
  \vspace{2ex}
\end{subfigure}
\begin{subfigure}{15.0cm}
  \centering
  \includegraphics[width=15.0cm,left]{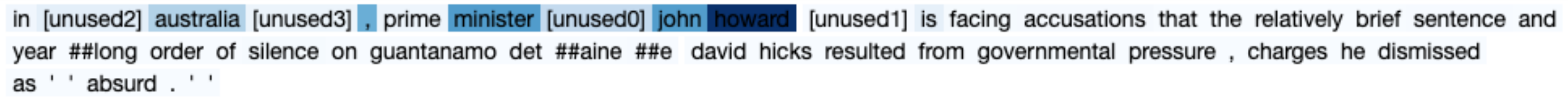}
  \caption{}
  \label{fig:nytatt}
\end{subfigure}
\caption{Examples of $A^{s2t}$ in the cross-stitch component, the comparative contribution of each token is visualized by the blue level, where the higher the blue the bigger the contribution. Figure~\ref{fig:bioatt} (above) shows the $A^{s2t}$ visualization for predicting the masked token in $(e_1, \text{\mask}, e_2)$, where the ground truth relation is \emph{may\_treat}, and Figure~\ref{fig:nytatt} (below) does $A^{s2t}$ visualization for predicting the KG relation \emph{/people/person/nationality}.}
\label{fig:caseatt}
\end{figure*}

\subsection{Cross-stitch Gate Weights vs. Noise}\label{ap:noise}

In order to analyze how the XBE model dynamically controls information flow between the encoders, we construct several sets of synthesized sentence bags differing in the proportion of noisy sentences they contain, similarly to \citet{hu2021knowledge}.
Specifically, given a target entity pair $(e_1, e_2)$ we create a synthesized sentence bag in which each valid sentence is created by the combination of the entity pair and the context that expresses their relation, and each noisy one by randomly selecting a context representing a different relation.
This process is illustrated in Figure~\ref{fig:synsent}.
We use the NYT10m dataset~\cite{gao2021manual}, which is a manually annotated version of the NYT10 test set, as data source and create $6$ sets of synthesized sentence bags with noise settings varying from $5/30$ to $30/30$, where $5/30$ ($30/30$) denotes that in the set, each bag has $30$ sentences and contains $5$ ($30$) noisy sentences.
Each set contains about $4$k entity pairs and each entity pair has $30$ sentences, for a total of about $130$k sentences.

We train one XBE model on each of the six sets with varying noise proportions and observe the gate weights of the cross-stitch mechanism, $\mathbf{G}_i^{t2s}$ in Equation~\ref{eq:gates}, which control the amount of information that flows into the next layer of text encoder. We show the weights with respect to different noise ratios in Figure~\ref{fig:noisegate}.
From the Figure~\ref{fig:noisegate}, we can observe that the gate weights (i.e., $\mathbf{G}_i^{t2s}$) tend to decrease as the noise ratio increases, indicating that the proposed cross-stitch mechanism of XBE effectively filters out noisy sentences and thereby aids the text encoder in extracting effective features. This observation is a possible explanation for the performance gain from the cross-stitch mechanism found in the ablation study (Table~\ref{tab:ablation}).

\subsection{Qualitative Examples}
\label{sec:casestudy}

We provide a few qualitative examples intended to demonstrate how the proposed cross-stitch mechanism can impact the performance of DS-RE, which are shown in Table~\ref{tab:casestudy}.
We can observe that the cross-stitch mechanism appears to facilitate DS-RE especially when a sentence bag is noisy. For instance, although the bag B1 fails to describe the \emph{/people/person/place\_lived} relation, the proposed model can utilize useful information from KG through cross-stitch and thus correctly predicts the relation.
Similarly, 
%since X-stitch realizes the encoder level communication between text an KG, 
the the model can correctly a identify \emph{may\_be\_treated\_by} relation from the bag B4, which does not explicitly describe the target relation. Please see the supplementary material for further results. 

We also visualize cross-attention weights $A^{s2t}$, which indicate the attention values over textual tokens used by the KG encoder to construct hidden representations.
As shown in Figure~\ref{fig:bioatt}, for the representation of the KG relation token \emph{may\_treat}, the cross-stitch mechanism assigns higher attention score on informative tokens such as ``drug'' and ``treating'' than the irrelevant ones from``more than 20 years''. 
Similarly, as shown in Figure~\ref{fig:nytatt}, in order to encode \emph{/people/person/nationality}, the cross-stitch mechanism focuses on the token ``minister'', which implicitly conveys the meaning of nationality, 
%(i.e., if ``X is a minister of a country'', then ``X has the nationality of the country''), 
than irrelevant tokens such as ``facing''.

\section{Additional Related Work}
%One drawback of traditional supervised RE paradigm is the requirement of a large amount of annotated data, which is expensive and time-consuming. To address this issue, DS is proposed by ~\cite{mintz2009distant} to generate large amounts of training data with limited supervision, but always suffers from wrong labelling problem. 

% many other multi-task information sharing architectures \cite{ruder2019latent}, but cross-stitch is simple and can easily combined with pretrained transformers, need to explore different architectures in future work

%Prior work has tackled noisy sentences in two ways.
% One approach is to extract relations not from a single sentence, but from multiple sentences mentioning the same entity pair.
% Recent RE models taking this approach are equipped with an attention mechanism; the idea being that the model should learn to attend more to informative sentences and less to noisy sentences \cite{lin2016neural,ji2017distant,du2018multi,jat2018improving,han2018neural,han2018hierarchical,dai2021two}.
% A drawback of cross-sentence attention is that it is not applicable to the many cases in which DS yields only a single sentence for a given entity pair.

In this section we discuss related work besides the approaches already mentioned in the introduction.
To improve the performance of a DS-RE model, recently, researchers introduce various attention mechanisms. \newcite{lin2016neural} propose a relation vector based attention mechanism. \newcite{jat2018improving,du2018multi} propose multi-level (e.g., word-level and sentence-level) structured attention mechanism. \newcite{ye2019distant} apply both intra-bag and inter-bag attention for DS-RE. \newcite{han2018hierarchical} propose a relation hierarchy based attention mechanism. \newcite{jia2019arnor} propose an attention regularization framework for DS-RE. To handle the one-instance sentence bags, \newcite{li2020self} propose a new selective gated mechanism.  
%However, these models rely only on noisy sentence evidences from DS, neglecting complementary information from KG.

%Besides textual evidence from DS, researchers also leverage external evidence for DS-RE. 
%\newcite{roller2015improving} use the inference pattern learned from KG for eliminating potentially related entity pairs from negative samples. \newcite{liu2017soft} propose a soft-label method that generates new training instances by dynamically correct wrong labels of training instances. 
\newcite{ji2017distant} apply entity descriptions generated from Freebase and Wikipedia as extra evidences, \newcite{lin2017neural} utilize multilingual text as extra evidences and \newcite{vashishth2018reside} use multiple side information including entity types, dependency and relation alias information for DS-RE. \newcite{alt2019fine} utilize pre-trained language model for DS-RE. \newcite{sun2019leveraging} apply relational table extracted from Web as extra evidences for DS-RE. \newcite{zeng2017incorporating} apply two-hop KG paths
%(e.g., \emph{/location/birthplace}, \emph{location/contains}) 
%identified from two-hop textual paths as extra evidences 
for DS-RE. \newcite{dai2021two} introduce multi-hop paths over a KG-text joint graph for DS-RE.

KG has been proved to be effective for DS-RE.
\newcite{han2018neural} propose a joint model that adopts a KG embeddings based attention mechanism.
\newcite{dai2019incorporating} extend the framework of \newcite{han2018neural} by introducing multiple KG paths as extra evidences for DS-RE. \newcite{hu2019improving} propose a multi-layer attention-based framework to utilize both KG and textual signals for DS-RE. Based on the extensive analysis about the effect of KG and attention mechanism on DS-RE, \newcite{hu2021knowledge} proposed a straightforward but strong model and achieve a significant performance gain. However these methods mostly employ shallow integration of KG and text such as representations concatenation and KG embedding based attention mechanism. To fully take advantage of KG for DS-RE, in this paper, we propose a novel model to realize deep encoder level integration of KG and text.

\section{Limitations}
We focus only on one particular NLP task (i.e., DS-RE) to explore the effective way to jointly encoding KG and text, and thus further work is required to determine to what extend the proposed XBE can be generalized into multiple NLP tasks. 
Therefore, our work carries the limitation that the performance gain in DS-RE does not guarantee that it is effective in other NLP tasks such as Knowledge Graph Completion and Question Answering, where the combination of KG and text is needed. For this reason, we empathize the importance of multi-tasking for exploring such research question. In addition, we only utilize monolingual datasets to conduct evaluation and thus further work is required to investigate the effectiveness of the proposed model on multi-lingual datasets.

\section{Conclusions and Future Work}
We proposed a cross-stitch bi-encoder architecture, XBE, to leverage the complementary relation between KG and text for distantly supervised relation extraction. 
%Specifically, the proposed model consists of two parallel encoders for KG and text respectively and the interaction is realize by a gated mechanism call CRST. CRST allows the information flow between certain layers of the encoders for mutual learning target-relevant features and filtering out noisy information. 
Experimental results on both Medline21 and NYT10 datasets prove the robustness of our model because the proposed model achieves significant and consistent improvement as compared with strong baselines and achieve a new state-of-the-art result on the widely used NYT10 dataset.
Possible future work includes a more thorough investigation of how communication between KG encoder and text encoder influences the performance, as well as a more complex KG encoder that can not only handle relation triples, but arbitrary KG subgraphs, which could have applications in, e.g., multi-hop relation extraction.

\section*{Acknowledgements}
This work was supported by JST CREST Grant Number JPMJCR20D2 and JSPS KAKENHI Grant Number 21K17814. We are grateful to the anonymous reviewers for their constructive comments.

% Entries for the entire Anthology, followed by custom entries
\bibliography{emnlp2022}
\bibliographystyle{acl_natbib}

\appendix
\section{Appendix}

\subsection{Cross-stitch Layer Selection}

\begin{figure}[t]
\centering
\includegraphics[width=\columnwidth]{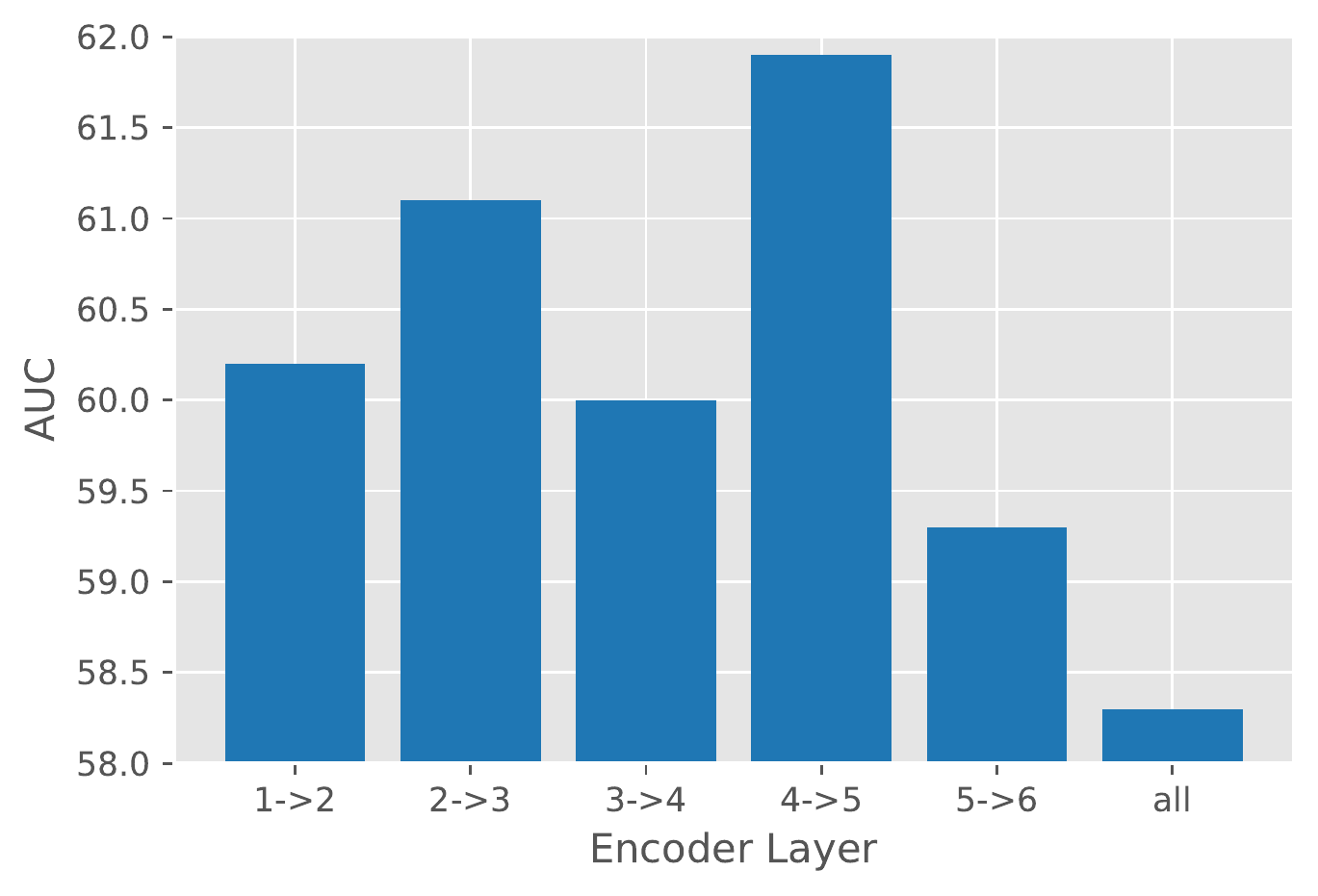}
\caption{Interacting layer wise performance on a development set of Medline21, where, for instance ``4->5'' means locating cross-stitch between Layer4 and Layer5, as shown in Figure~\ref{fig:introxbe}, ``all'' means setting cross-stitch between all adjacent layers.}
\label{fig:layerauc}
\end{figure}

Since the first few layers of BERT are the basis for the high level semantic task~\cite{jawahar2019does}, we place the cross-stitch in the layers $1 \sim 6$, and conduct a layer by layer analysis to find the best fitting layers in a development set. The development set is obtained by $30\%$ random selection from the training set of the Medline21. The layer-pair wise performance is shown in Figure~\ref{fig:layerauc}, which indicates that setting cross-stitch between Layer4 and Layer5 achieves better AUC than the others, which might be because the information encoded by Layer4 is complementary. In addition, ``all'' fails to outperform the others, which might be because not all the layers are complementary, for instance the KG encoder provides very little syntactic information for the text encoder.
\begin{table*}[t]
\centering
\scalebox{0.8}{
\begin{tabular}{c c c c c c c c}
\hline
\textbf{System} & AUC & P@0.1k & P@0.2k & P@0.3k & P@0.5k & P@1k & P@2k\\\hline\hline
XBE (Fixed Gate) & 68.82 & 98.0 & \textbf{96.5} & 94.0 & 92.4 & 84.5 & 62.2 \\ \hline
XBE (Dynamic Gate) & \textbf{70.50} & \textbf{99.0} & 96.0 & \textbf{95.6} & \textbf{94.4} & \textbf{85.8} & \textbf{63.2} \\ \hline
\end{tabular}
}
\caption{P@N and AUC from XBE with Fixed and Dynamic Gate on NYT10 dataset.}
\label{tab:gatenyt}
\end{table*}
\begin{table*}[t]
\centering
\scalebox{0.8}{
\begin{tabular}{c c c c c c c c}
\hline
\textbf{System} & AUC & P@0.1k & P@0.2k & P@0.3k & P@0.5k & P@1k & P@2k\\\hline\hline
XBE (Fixed Gate) & 61.87 & 93.0 & 89.5 & 88.7 & 86.2 & \textbf{76.6} & \textbf{56.4} \\ \hline
XBE (Dynamic Gate) & \textbf{61.88} & \textbf{94.0} & \textbf{91.0} & \textbf{89.3} & \textbf{86.4} & 76.1 & 56.1 \\ \hline
\end{tabular}
}
\caption{P@N and AUC from XBE with Fixed and Dynamic Gate on Medline21.}
\label{tab:gatebio}
\end{table*}
\begin{table}[h]
\centering
\scalebox{0.9}{
\begin{tabular}{lcccc}
\hline
&\multicolumn{2}{c}{\textbf{Medline21}} &\multicolumn{2}{c}{\textbf{NYT10}}\\\cmidrule(lr){2-3}\cmidrule(lr){4-5}
\textbf{Model} & AUC & P@2k & AUC & P@2k\\\cmidrule(lr){1-3}\cmidrule(lr){4-5}
BRE+CE & 55.3 & 53.8 & 63.2 & 58.7 \\\cmidrule(lr){1-3}\cmidrule(lr){4-5}
\ours & \textbf{61.9} & 56.1 & \textbf{70.5} & \textbf{63.2} \\\cmidrule(lr){1-3}\cmidrule(lr){4-5}
\makecell[r]{- $\mathbf{r}_{ht}$} & 59.02 & \textbf{56.9} & 68.64 & 62.1
\\\hline
\end{tabular}
}
\caption{P@N and AUC from XBE with removed $\mathbf{r}_{ht}$.}
\label{tab:ablationrht}
\end{table}
\subsection{Dynamic Gate vs. Fixed Gate}\label{ap:gate}
Two strategies can be applied to calculate the weights for  $\mathbf{G}_i^{t2s}$ and $\mathbf{G}_i^{s2t}$ in Figure~\ref{fig:crst}: one is using fixed weights of gate throughout the entire training process; another is the proposed dynamic control of weights evaluated via Equation~\ref{eq:gates} and Equation~\ref{eq:gatet} respectively. Table~\ref{tab:gatenyt} and Table~\ref{tab:gatebio} show the performance comparison between the fixed gate and the proposed dynamic gate. We set the value of the fixed gate as $0.5$ in this work. The results show that our proposed dynamic gate achieves better performance than the fixed gate, indicating the effectiveness of the proposed XBE model on dynamically controlling information flow from one layer to the next.

\subsection{Impact of $\mathbf{r}_{ht}$}\label{ap:rht}
We conduct ablation study to detect the impact of $\mathbf{r}_{ht}$ in \secref{sec:model-re} on the overall performance. The results are shown in Table~\ref{tab:ablationrht}, where ``- $\mathbf{r}_{ht}$'' denotes the XBE model without $\mathbf{r}_{ht}$. We can observe that the performance slightly degrades without $\mathbf{r}_{ht}$, indicating that $\mathbf{r}_{ht}$ has limited contribution to the performance gain comparing with other components of the XBE model.

%supplementary material
\subsection{Supplementary Material}
\begin{table*}[t]
\centering
\begin{tabular}{|l|c|}
\hline
\textbf{Hyperparameter} & \textbf{Value} \\\hline
word embedding dimension & 50 \\ \hline
KG embedding dimension & 50 \\ \hline
position embedding dimension & 5 \\ \hline
CNN window size & 3 \\ \hline
CNN filter number &  230 \\ \hline
dropout rate & 0.5 \\ \hline
learning rate & 0.01 \\ \hline
batch size & 160 \\ \hline
\end{tabular}
\caption{Hyperparameters used in PCNN+ATT on Medline21 dataset. The experiments are conducted on Nvidia Titan X(Pascal) GPU.}
\label{tab:hypcnn}
\end{table*}
\begin{table*}
\centering
\begin{tabular}{|l|c|}
\hline
\textbf{Hyperparameter} & \textbf{Value} \\\hline
word embedding dimension & 50 \\ \hline
KG embedding dimension & 50 \\ \hline
position embedding dimension & 5 \\ \hline
CNN window size & 3 \\ \hline
CNN filter number &  100 \\ \hline
dropout rate & 0.5 \\ \hline
learning rate (for sentences) & 0.02 \\ \hline
learning rate (for KG) & 0.001 \\ \hline
batch size & 100 \\ \hline
\end{tabular}
\caption{Hyperparameters used in JointE on Medline21 dataset. The experiments are conducted on Nvidia Titan X(Pascal) GPU.}
\label{tab:hyjointe}
\end{table*}
\begin{table*}
\centering
\begin{tabular}{|l|c|}
\hline
\textbf{Hyperparameter} & \textbf{Value} \\\hline
word embedding dimension & 50 \\ \hline
KG embedding dimension & 50 \\ \hline
position embedding dimension & 5 \\ \hline
CNN window size & 3 \\ \hline
CNN filter number &  100 \\ \hline
dropout rate & 0.5 \\ \hline
learning rate (for sentences) & 0.1 \\ \hline
learning rate (for KG) & 0.001 \\ \hline
batch size & 100 \\ \hline
\end{tabular}
\caption{Hyperparameters used in RELE on Medline21 dataset. The experiments are conducted on Nvidia Titan X(Pascal) GPU.}
\label{tab:hyrele}
\end{table*}
\begin{table*}
\centering
\begin{tabular}{|l|c|}
\hline
\textbf{Hyperparameter} & \textbf{Value} \\\hline
learning rate & 3e-5 \\ \hline
hidden size & 768 \\ \hline
weight decay rate & 1e-5 \\ \hline
Adam epsilon & 1-e8 \\ \hline
warmup steps & 500 \\ \hline
batch size & 100 \\ \hline
maximum epochs & 15 \\ \hline
\end{tabular}
\caption{Hyperparameters used in BRE+CE on Medline21 dataset. The experiments are conducted on a NVIDIA GeForce GTX 1080 TI GPU.}
\label{tab:hybre}
\end{table*}
\begin{table*}
\centering
\begin{tabular}{|l|c|c|}
\hline
\textbf{Hyperparameter} & \textbf{Value} (Medline21) & \textbf{Value} (NYT10) \\\hline
learning rate & 3e-5 & 3e-5 \\ \hline
hidden size & 768 & 768 \\ \hline
weight decay rate & 1e-5 & 1e-5 \\ \hline
Adam epsilon & 1-e8 & 1-e8 \\ \hline
warmup steps & 500 & 500 \\ \hline
batch size & 100 & 80 \\ \hline
maximum epochs & 15 & 10 \\ \hline
\end{tabular}
\caption{Hyperparameters used in CRE+KA model. The experiments are conducted on a NVIDIA GeForce GTX 1080 TI GPU.}
\label{tab:hycrst}
\end{table*}
\begin{table*}
\centering
\begin{tabular}{|l|c|c|}
\hline
\textbf{Hyperparameter} & \textbf{Value} (Medline21) & \textbf{Value} (NYT10) \\\hline
learning rate & 3e-5 & 3e-5 \\ \hline
hidden size & 768 & 768 \\ \hline
weight decay rate & 1e-5 & 1e-5 \\ \hline
Adam epsilon & 1-e8 & 1-e8 \\ \hline
warmup steps & 500 & 500 \\ \hline
batch size & 100 & 80 \\ \hline
$w$ & 1.0 & 0.6 \\ \hline
$\lambda_t$ & 1.0 & 1.0 \\ \hline
$\lambda_s$ & 1-e4 & 1-e4 \\ \hline
maximum epochs & 15 & 10 \\ \hline
\end{tabular}
\caption{Hyperparameters used in our proposed XBE model. The experiments (Medline21) are conducted on Nvidia Titan X(Pascal) GPU, and the experiments (NYT10) are conducted on a NVIDIA GeForce GTX 1080 TI GPU.}
\label{tab:hycrst}
\end{table*}
\begin{table*}
\centering
\scalebox{0.8}{
\begin{tabular}{c|l|l|c|c|c}
\hline
Bag &\textbf{Sentence} & Target Relation &\textbf{XBE}& \makecell{XBE \\ - X-stitch} & BRE+CE \\
\hline
\hline
B1 & \makecell[l]{ ... ' idol ' finalist arrested \textbf{jessica sierra} , one of the \\ top 10 finalists on '' american idol '' in 2005 , \\ was arrested early yesterday in \textbf{tampa}, ...} & \makecell{/people/person/\\place\_lived} & \cmark & \xmark & \xmark\\ \hline
B2 & \makecell[l]{... accepted a job , a friend suggested he check out \\ a \textbf{san francisco} start-up , \textbf{powerset} , which, ...} & \makecell{/business/company/\\place\_founded} & \cmark & \xmark & \xmark\\ \hline
B3 & \makecell[l]{..., the coordinator of aurore , a renewable energy \\ service company in \textbf{auroville} , \textbf{india} .} & \makecell{/location/location/\\contains} & \cmark & \xmark & \xmark\\ \hline
B4 & \makecell[l]{president \textbf{gloria macapagal-arroyo} in a statement \\ released after she voted in her province of \textbf{pampanga} , \\ north of manila , said the country ...} & \makecell{/people/person/\\place\_lived} & \cmark & \xmark & \xmark\\ \hline
B5 & \makecell[l]{..., a track architect known for his work on \\ tracks in bahrain , shanghai and \textbf{sepang} , \textbf{malaysia} .} & \makecell{/location/location/\\contains} & \cmark & \xmark & \xmark\\ \hline\hline
B6 & \makecell[l]{In the clinical context hair analysis \\ be advantageously used to monitor the abuse of \\ analgesic combinations with \textbf{C0054234\#ent} , \\ common among \textbf{C0018681\#ent} patients .} & \makecell{may\_treat} & \cmark & \xmark & \xmark\\ \hline
B7 & \makecell[l]{The solute carrier family 45 a3 member \\ ( \textbf{C1822765\#ent} ) , known also as \textbf{C0965053\#ent}, \\ has been implicated with ...} & \makecell{gene\_encodes\_\\gene\_product} & \cmark & \xmark & \xmark\\ \hline
B8 & \makecell[l]{... and investigated the effects of \textbf{C1098320\#ent} \\ on ID remodelling during development \\ of \textbf{C0018801\#ent} } & \makecell{may\_treat} & \cmark & \xmark & \xmark\\ \hline
B9 & \makecell[l]{\textbf{C1832024\#ent} interferes with anaerobic glycolysis \\ pentose cycle \textbf{C0035547\#ent}.} & \makecell{chemical\_or\_drug\_\\affects\_gene\_product} & \cmark & \xmark & \xmark\\ \hline
B10 & \makecell[l]{... to minimize acute and \textbf{C0030201\#ent} \\ has been provided by microspheres that slowly \\ release \textbf{C0006400\#ent} ( MS-Bup ) ...} & \makecell{may\_be\_\\treated\_by} & \cmark & \xmark & \xmark\\ \hline
\end{tabular}
}
\caption{Some practical results, where each bag contains one sentence, \cmark (or \xmark) represents the correct (or incorrect) prediction of the target relation. We can observe that the proposed X-stitch can facilitate XBE to correctly identify the relation especially when a sentence bag is noisy or implicitly represents the target relation.}
\label{tab:casestudy}
\end{table*}
\end{document}